\documentclass[conference]{IEEEtran}

\usepackage{amsmath,amssymb,amsfonts}
\usepackage{graphicx}
\usepackage{xcolor}

\usepackage{mathptmx} %

\usepackage{fancyhdr}
\usepackage{footmisc}
\usepackage[normalem]{ulem}
\usepackage[hyphens]{url}
\usepackage[sort,nocompress]{cite}
\usepackage[final]{microtype}
\usepackage{balance}

\usepackage[utf8]{inputenc}
\usepackage{amsmath}
\usepackage{amsfonts}
\usepackage{amssymb}
\usepackage{graphicx}
\usepackage[setmargin=true,hide=true,marginparwidth=0.4in]{marginalia}
\usepackage[numbers]{natbib}
\usepackage{xspace}
\usepackage{newtxtext, newtxmath}
\usepackage{booktabs}
\usepackage{multirow}
\usepackage{listings}
\usepackage[framemethod=tikz]{mdframed}
\usepackage{colortbl}
\usepackage{subcaption}
\usepackage{tabularx}

\usepackage{algorithm}
\usepackage{algpseudocode}

\usepackage[bookmarks=true,breaklinks=true,letterpaper=true,colorlinks,linkcolor=black,citecolor=blue,urlcolor=black]{hyperref}
\usepackage{cleveref}
\usepackage{etoolbox}

\makeatletter
\pretocmd{\NAT@citexnum}{\@ifnum{\NAT@ctype>\z@}{\let\NAT@hyper@\relax}{}}{}{}
\makeatother

\makeatletter
\newcommand{\paragraph@gobblepar}[1]{%
  \@ifnextchar\par{\paragraph@gobblepar}{}%
}
\renewcommand{\paragraph}[1]{%
  \par\bigskip\noindent\textbf{#1.}\;
  \@ifnextchar\par{\paragraph@gobblepar}{}%
}
\makeatother

\DeclareMathOperator*{\argmin}{arg\,min}

\let\originalleft\left
\let\originalright\right
\renewcommand{\left}{\mathopen{}\mathclose\bgroup\originalleft}
  \renewcommand{\right}{\aftergroup\egroup\originalright}

\newcommand{\mca}{llvm-mca\xspace}
\newcommand{\exegesis}{llvm\_sim\xspace}
\newcommand{\scaffold}{surrogate\xspace}

\newcommand{\Scaffold}{Surrogate\xspace}
\newcommand{\Scaffolds}{Surrogates\xspace}
\newcommand{\latency}{\texttt{WriteLatency}\xspace}
\newcommand{\portmap}{\texttt{PortMap}\xspace}
\newcommand{\nummicroops}{\texttt{NumMicroOps}\xspace}
\newcommand{\dispatchwidth}{\texttt{DispatchWidth}\xspace}
\newcommand{\reorderbuffersize}{\texttt{ReorderBufferSize}\xspace}
\newcommand{\readadvance}{\texttt{ReadAdvanceCycles}\xspace}
\newcommand{\name}{DiffTune\xspace}
\newcommand{\nops}{837\xspace}
\newcommand{\nparams}{11265\xspace} %
\newcommand{\psize}{$10^{19336}$\xspace} %

\newcommand{\instr}[1]{{\color[HTML]{0600F5}\ttfamily #1}}

\lstdefinelanguage
   [x64]{Assembler}     %
   [x86masm]{Assembler} %
   {
     alsoletter={\%},
     otherkeywords={\%rbx, \%r8d, \%r13d, \%eax, \%rsp},
     ndkeywords={xorl, pushq, testl, addl, shrq, xorq, xor},
}

\begin{document}
\title{\name{}: Optimizing CPU Simulator Parameters with Learned Differentiable \Scaffolds}

\author{\IEEEauthorblockN{Alex Renda}
\IEEEauthorblockA{\textit{MIT CSAIL} \\
renda@csail.mit.edu}
\and
\IEEEauthorblockN{Yishen Chen}
\IEEEauthorblockA{\textit{MIT CSAIL} \\
ychen306@mit.edu}
\and
\IEEEauthorblockN{Charith Mendis}
\IEEEauthorblockA{\textit{MIT CSAIL} \\
charithm@mit.edu}
\and
\IEEEauthorblockN{Michael Carbin}
\IEEEauthorblockA{\textit{MIT CSAIL} \\
mcarbin@csail.mit.edu}
}

\maketitle

\begin{abstract}
  CPU simulators are useful tools for modeling CPU execution behavior.
However, they suffer from inaccuracies due to the cost and complexity of setting their fine-grained parameters, such as the latencies of individual instructions.
This complexity arises from the expertise required to design benchmarks and measurement frameworks that can precisely measure the values of parameters at such fine granularity.
In some cases, these parameters do not necessarily have a physical realization and are therefore fundamentally approximate, or even unmeasurable.

In this paper we present \name{}, a system for learning the parameters of x86 basic block CPU simulators from coarse-grained end-to-end measurements.
Given a simulator, \name{} learns its parameters by first replacing the original simulator with a \emph{differentiable surrogate}, another function that approximates the original function;
by making the surrogate differentiable, \name{} is then able to apply gradient-based optimization techniques even when the original function is non-differentiable, such as is the case with CPU simulators.
With this differentiable surrogate, \name{} then applies gradient-based optimization to produce values of the simulator's parameters that minimize the simulator's error on a dataset of ground truth end-to-end performance measurements.
Finally, the learned parameters are plugged back into the~original~simulator.

\name{} is able to automatically learn the entire set of microarchitecture-specific parameters within the Intel x86 simulation model of \mca{}, a basic block CPU simulator based on LLVM's instruction scheduling model.
\name{}'s learned parameters lead \mca to an average error that not only matches but lowers that of its original, expert-provided~parameter~values.
\end{abstract}

\section{Introduction}
\label{sec:introduction}

Simulators are widely used for architecture research to model the interactions of architectural components of a system~\citep{binkert_gem5_2011,llvm-mca,iaca,ptlsim,marss,sanchez_zsim_2013}.
For example, \emph{CPU simulators}, such as \mca{}~\citep{llvm-mca}, and \exegesis{}~\citep{exegesis}, model the execution of a processor at various levels of detail, potentially including abstract models of common processor design concepts such as dispatch, execute, and retire stages~\citep{patterson_hennessy}.
CPU simulators can operate at different granularities, from analyzing just \emph{basic blocks}, straight-line sequences of assembly code instructions, to analyzing whole programs.
Such simulators allow performance engineers to reason about the execution behavior and bottlenecks of programs run on a given processor.

However, precisely simulating a modern CPU is challenging: not only are modern processors large and complex, but many of their implementation details are proprietary, undocumented, or only loosely specified even given the thousands of pages of vendor-provided documentation that describe any given processor.
As a result, CPU simulators are often composed of coarse abstract models of a subset of processor design concepts.
Moreover, each constituent model typically relies on a number of approximate design parameters, such as the number of cycles it takes for an instruction to pass through the processor's execute stage.
Choosing an appropriate level of model detail for simulation, as well as setting simulation parameters, requires significant expertise.
In this paper, we consider the challenge of setting the parameters of a CPU simulator given a fixed level of model detail.

\paragraph{Measurement}
One methodology for setting the parameters of such a CPU simulator is to gather fine-grained measurements of each individual parameter's realization in the physical machine~\citep{agner,abel_uops_2019} and then set the parameters to their measured values~\citep{mca-agner,mca-uops}.
When the semantics of the simulator and the semantics of the measurement methodology coincide, then these measurements can serve as effective parameter values.
However, if there is a mismatch between the simulator and the measurement methodology, then measurements may not provide effective parameter settings~\citep[Section~5.2]{ritter_pmevo_2020}.
Moreover, some parameters may not correspond to measurable values.

\paragraph{Optimizing simulator parameters}
An alternative to developing detailed measurement methodologies for individual parameters is to infer the parameters from coarse-grained end-to-end measurements of the performance of the physical machine~\citep{ritter_pmevo_2020}. 
Specifically, given a dataset of benchmarks, each labeled with their true behavior on a given CPU (e.g., with their execution time or with microarchitectural events, such as cache misses), identify a set of parameters that minimize the error between the simulator's predictions and the machine's true behavior.
This is generally a discrete, non-convex optimization problem for which classic strategies, such as random search~\citep{ansel_opentuner_2014}, are intractable because of the size of the parameter space (approximately \psize possible parameter settings in one simulator, \mca{}).

\paragraph{Our approach: \name{}}

In this paper, we present \name{}, an optimization algorithm and implementation for learning the parameters of programs.
We use \name{} to learn the parameters of x86 basic block CPU simulators.

\name's algorithm takes as input a program, a description of the program's parameters, and a dataset of input-output examples describing the program's desired output, then produces a setting of the program's parameters that \NA{minimizes} the discrepancy between the program's actual and desired output.
The learned parameters are then plugged back into the~original~program.

The algorithm solves this optimization problem via a \emph{differentiable surrogate} for the program~\citep{queipo_surrogate_2005,shirobokov_differentiating_2020,louppe_adversarial_2017,grathwohl_backpropagation_2018,she_neuzz_2019}.
A surrogate is an approximation of the function from the program's parameters to the program's output.
By requiring the surrogate to be differentiable, it is then possible to compute the surrogate's gradient and apply gradient-based optimization~\citep{robbins_stochastic_1951,kingma_adam_2014} to identify a setting of the program's parameters that minimize the error between the program's output (as predicted by the surrogate) and the desired output. %

To apply this to basic block CPU simulators, we instantiate \name's surrogate with a neural network that can mimic the behavior of a simulator.
This neural network takes the original simulator input (e.g., a sequence of assembly instructions) and a set of proposed simulator parameters (e.g., dispatch width or instruction latency) as input, and produces the output that the simulator would produce if it were instantiated with the given simulator's parameters.
We derive the neural network architecture of our surrogate from that of Ithemal~\citep{mendis_ithemal_2019}, a basic block throughput estimation neural network.

\paragraph{Results}
Using \name{}, we are able to learn the entire set of \nparams microarchitecture-specific parameters in the Intel x86 simulation model of \mca{}~\citep{llvm-mca}.
\mca is a CPU simulator that predicts the execution time of basic blocks.
\mca{} models instruction dispatch, register renaming, out-of-order execution with a reorder buffer, instruction scheduling based on use-def latencies, execution by dispatching to ports, a load/store unit ensuring memory consistency, and a retire~control~unit.\footnote{We note that \mca{} does not model the memory hierarchy.}

We evaluate \name{} on four different x86 microarchitectures, including both Intel and AMD chips.
Using only end-to-end supervision of the execution time measured per-microarchitecture of a large dataset of basic blocks from \citet{chen_bhive_2019}, we are able to learn parameters from scratch that lead \mca{} to have an average error of $24.6\%$, down from an average error of $30.0\%$ with \mca's expert-provided parameters. In contrast, black-box global optimization with OpenTuner~\citep{ansel_opentuner_2014} is unable to identify parameters with less~than~$100\%$~error.

\paragraph{Contributions} We present the following contributions:

\begin{itemize}
\item We present \name, an algorithm for learning ordinal parameters of programs from input-output examples.
\item We present an implementation of \name for x86 basic block CPU simulators that uses a variant of the Ithemal model as a differentiable surrogate.
\item We evaluate \name{} on \mca{} and demonstrate that \name{} can learn the entire set of microarchitectual parameters in \mca{}’s Intel x86 simulation model.
\item \NA{We present case studies of specific parameters learned by \name{}. Our analysis demonstrates cases in which \name learns semantically correct parameters that enable \mca to make more accurate predictions. Our analysis also demonstrates cases in which \name learns parameters that lead to higher accuracy but do not correspond to reasonable physical values on the CPU.}
\end{itemize}

Our results show that \name{} offers the promise of a generic, scalable methodology to learn detailed performance models with only end-to-end measurements, reducing performance optimization tasks to simply that of gathering data.

\section{Background: Simulators}
\label{sec:background}

Simulators comprise a large set of tools for modeling the execution behavior of computing systems, at all different levels of abstraction: from cycle-accurate simulators to high-level cost models.
These simulators are used for a variety of applications:
\begin{itemize}
\item gem5~\citep{binkert_gem5_2011} is a detailed, extensible full system simulator that is frequently used for computer architecture research, to model the interaction of new or modified components with the rest of a CPU and memory system.
\item IACA~\citep{iaca} is a static analysis tool released by Intel that models the behavior of modern Intel processors, including undocumented Intel CPU features, predicting code performance.
  IACA is used by performance engineers to diagnose and fix bottlenecks in hand-engineered code~snippets~\citep{laukemann-osaca}.
\item LLVM~\citep{llvm} includes internal CPU simulators to predict the performance of generated code~\citep{pohl_vectorization_2020,mendis_goslp_2018}.
  LLVM uses these CPU simulators to search through the code optimization space, to generate more optimal code.
\end{itemize}

Though these simulators are all simplifications of the true execution behavior of physical systems, they are still highly complex pieces of software.

\subsection{\mca{}}

For example, consider \mca{}~\citep{llvm-mca}, an out-of-order superscalar CPU simulator included in the LLVM~\citep{llvm} compiler infrastructure.
The main design goal of \mca{} is to expose LLVM's instruction scheduling model for testing.
\mca{} takes \emph{basic blocks} as input, sequences of straight-line assembly instructions with no branches, jumps, or loops.
For a given input basic block, \mca{} predicts the timing of 100 repetitions of that block, measured in cycles.

\paragraph{Design}
\mca{} is structured as a generic, target-independent simulator parameterized on LLVM's internal model of the target hardware.
\mca{} makes two core modeling assumptions. First, it assumes that the simulated program is not bottlenecked by the processor frontend; in fact, \mca{} ignores instruction decoding entirely.
Second, \mca{} assumes that memory data is always in the L1 cache, and ignores the memory~hierarchy.

\mca{} simulates a processor in four main stages: \emph{dispatch}, \emph{issue}, \emph{execute}, and \emph{retire}.
The dispatch stage reserves physical resources (e.g., slots in the reorder buffer) for each instruction, based on the number of micro-ops the instruction is composed of.
Once dispatched, instructions wait in the issue stage until they are ready to be executed.
The issue stage blocks an instruction until its input operands are ready and until all of its required execution ports are available.
Once the instruction's operands and ports are available, the instruction enters the execute stage.
The execute stage reserves the instruction's execution ports and holds them for the durations specified by the instruction's port map assignment specification.
Finally, once an instruction has executed for its duration, it enters the retire stage.
In program order, the retire stage frees the physical resources that were acquired for each instruction.

\begin{figure*}
\centering
\includegraphics[width=\textwidth]{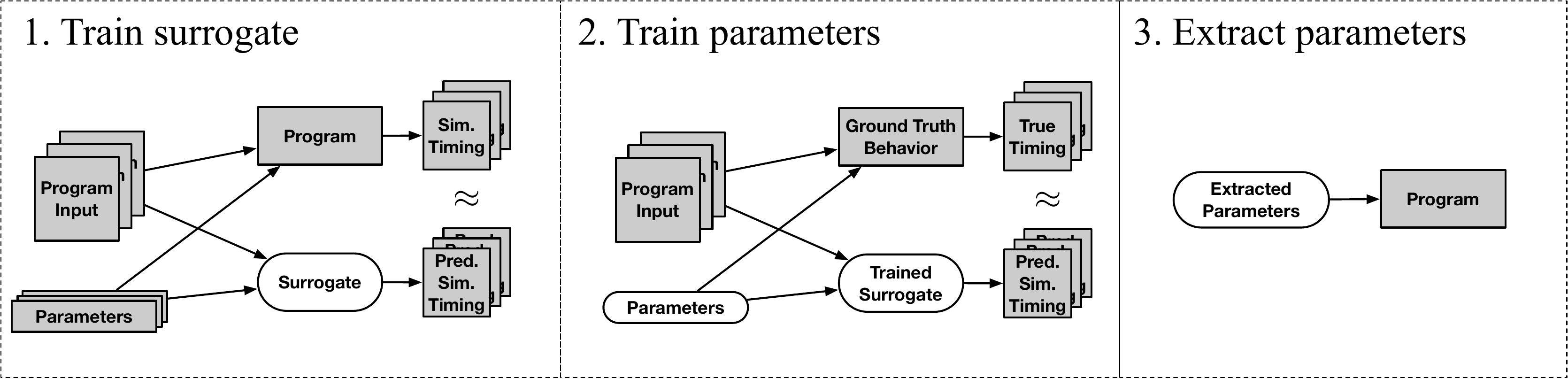}
\caption{\name block diagram.}
\label{fig:algorithm-diagram}
\end{figure*}

\begin{table*}
  \caption{Terms used in formalism in \Cref{sec:approach}.}
  \label{tab:terms}
  \centering
\begin{tabular}{cl}
  \toprule
  \textbf{Notation} & \textbf{Definition} \\
  \midrule
  $f$ & Program that we are trying to optimize. \\
  $\theta$ & Parameters of the program that we are trying to optimize. \\
  $x$ & Input to the program. \\
  $y$ & Output of the program on an input $x$. \\
  $\mathcal{I}$ & Dataset of program inputs to optimize against. \\
  $f^*$ & Ground truth behavior that we are trying to model with $f$. \\
  $\mathcal{L}$ & Loss function describing the error of a proposed solution. \\
  $\hat{f}$ & The \scaffold, which is trained to model the original program: $\hat{f} \approx f$. \\
  $D$ & Distribution that parameters $\theta$ are sampled from for training the \scaffold{}. \\
  \bottomrule
\end{tabular}
\end{table*}

\paragraph{Parameters} Each stage in \mca{}'s model requires parameters.
The \nummicroops parameter for each instruction specifies how many micro-ops the instruction is composed of.
The \dispatchwidth parameter specifies how many micro-ops can enter and exit the dispatch stage in each cycle.
The \reorderbuffersize parameter specifies how many micro-ops can reside in the issue and execute stages at the same time.
The \portmap parameters for each instruction specify the number of cycles for which the instruction occupies each execution port.
An additional \latency parameter for each instruction specifies the number of cycles before destination operands of that instruction can be read from, while \readadvance parameters for each instruction specify the number of cycles by which to accelerate the \latency of source operands (representing forwarding paths).

In sum, the $\nops$ instructions in our dataset (\Cref{sec:methodology}) lead to $\nparams$ total parameters with \psize possible configurations in \mca{}'s Haswell microarchitecture simulation.\footnote{\label{foot:psize}Based on \mca{}'s default, expert-provided values for these parameters, the $\nparams$ parameters induce a parameter space of \psize valid configurations; the actual values are only bounded by integer representation sizes.}

\subsection{Challenges}

These parameter tables are currently manually written for each microarchitecture, based on processor vendor documentation and manual timing of instructions.
Specifically, many of LLVM's \latency and \portmap parameters are drawn from the Intel optimization manual~\citep{intel-documentation,mca-intel}, Agner Fog's instruction tables~\citep{agner,mca-agner}, and uops.info~\citep{abel_uops_2019,mca-uops}, all of which contain latencies and port mappings for assembly instructions across different architectures and microarchitectures.

\paragraph{Measurability}
However, these documented and measured values do not directly correspond to parameters in \mca{}, because \mca{}'s parameters, and abstract simulator parameters more broadly, are not defined such that they have a single measurable value.
For instance, \mca{} defines exactly one \latency parameter per instruction.
However, \citet{agner} and \citet{abel_uops_2019} find that for instructions that produce multiple results in different destinations, the results might be available at different cycles.
Further, the latency for results to be available can depend on the actual value of the input operands.
Thus, there is no single measurable value that corresponds to \mca{}'s definition of \latency.

Different choices for how to map from measured latencies to \latency values result in different overall errors (as defined in \Cref{sec:methodology}).
For instance, when \mca{} is instantiated with \citet{abel_uops_2019}'s maximum observed latency for each instruction, \mca{} gets an error of $218\%$ when generating predictions for the Haswell microarchitecture; the median observed latency results in an error of $150\%$; and the minimum observed latency results in an error of $103\%$.

\section{Approach}
\label{sec:approach}

Tuning \mca{}'s $\nparams$ parameters among \psize valid configurations\footref{foot:psize} by exhaustive search is impractical.
Instead, we present \name{}, an algorithm for learning ordinal parameters of arbitrary programs from labeled input and output examples.
\name{} leverages learned differentiable surrogates to make the optimization process more tractable.

\paragraph{Formal problem statement}
Given a program $f : \Theta \to \mathcal{X} \to \mathcal{Y}$ parameterized on parameters $\theta : \Theta$, that takes inputs $x : \mathcal{X}$ to outputs $y : \mathcal{Y}$, and given a function $f^*: \mathcal{X} \to \mathcal{Y}$ that represents ground-truth behavior, find parameters $\theta \in \Theta$ to minimize the expected value of a cost function (called the loss function, representing error) $\mathcal{L} : \left(\mathcal{Y} \times \mathcal{Y}\right) \to \mathbb{R}_{\geq0}$ of the discrepancy between the behavior of the program $f$ and the ground-truth behavior $f^*$ on a dataset of program inputs $\mathcal{I} \subseteq \mathcal{X}$:
\begin{equation}
  \argmin_{\theta} \mathbb{E}_{x \sim \mathcal{I}} \left[ \mathcal{L}\left(f\left(\theta, x\right), f^*\left(x\right)\right) \right]
  \label{alg:problem}
\end{equation}

\paragraph{Algorithm} Figure~\ref{fig:algorithm-diagram} presents a diagram of our approach.
We first optimize the \scaffold $\hat{f} : \Theta \to \mathcal{X} \to \mathcal{Y}$ to mimic the original program, i.e., $\forall \theta,x. \hat{f}(\theta, x) \approx f(\theta, x)$.
Specifically, we optimize the \scaffold  to minimize the expectation of the loss $\mathcal{L}$ over program inputs from $\mathcal{I}$ and samples of $\theta$ from a pre-defined parameter sampling distribution~$D$:
\begin{equation}
  \argmin_{\hat{f}} \mathbb{E}_{x \sim \mathcal{I}, \theta \sim D}\left[\mathcal{L}\left(f\left(\theta, x\right), \hat{f}\left(\theta, x\right)\right)\right]
\label{alg:surrogate}
\end{equation}
We then optimize the parameters $\theta$ to minimize the discrepancy between predictions of the \scaffold{} $\hat{f}$ and the ground-truth behavior $f^*$.
Specifically, we find:
\begin{equation}
  \argmin_{\theta} \mathbb{E}_{x \sim \mathcal{I}} \left[\mathcal{L}\left(\hat{f}\left(\theta, x\right), f^*\left(x\right)\right)\right]
\label{alg:surrogate_problem}
\end{equation}
Finally, we extract the learned parameters $\theta$ from the optimization problem and plug them into the original program $f$, applying any constraints (e.g., that the parameters must be integers) that were not enforced when optimizing against~the~surrogate.

\paragraph{Discussion}
Note the similarity between \Cref{alg:problem} and \Cref{alg:surrogate_problem}: the two equations only differ by the use of $f$ and $\hat{f}$, respectively.
The close correspondence between forms makes clear that $\hat{f}$ stands in as a surrogate for the original program, $f$.
This is a general algorithmic approach~\citep{queipo_surrogate_2005} that is desirable when it is possible to choose $\hat{f}$ such that it is easier or more efficient to optimize $\theta$ using $\hat{f}$ than $f$.

\paragraph{Optimization}
In our approach, we choose $\hat{f}$ to be a neural network.
Neural networks are typically built as compositions of differentiable architectural components, such as \emph{embedding lookup tables}, which map discrete input elements to real-valued vectors; \emph{LSTMs}~\citep{hochreiter_lstm_1997}, which map input sequences of vectors to a single output vector; and \emph{fully connected layers}, which are linear transformations on input vectors.
By being composed of differentiable components, neural networks are end-to-end differentiable, so that they are able to be trained using gradient-based optimization.
Specifically, neural networks are typically optimized with stochastic first-order optimizations like stochastic gradient descent (SGD)~\citep{robbins_stochastic_1951}, which repeatedly calculates the network's error on a small sample of the training dataset and then updates the network's parameters in the opposite of the direction of the gradient to minimize the error.

By selecting a neural network as $\hat{f}$'s  representation, we are able to leverage $\hat{f}$'s differentiable nature not only to train $\hat{f}$ (solving the optimization problem posed in \Cref{alg:surrogate}) but also to solve the optimization problem posed in \Cref{alg:surrogate_problem} with gradient-based optimization.
This stands in contrast to $f$ which is, generally, non-differentiable and therefore does not permit the computation of its gradients.

\paragraph{Surrogate example}

A visual example of this is presented in Figure~\ref{fig:diff-approx-plot}, which shows an example of the timing predicted by \mca{} (blue) and a trained \scaffold of \mca{} (orange).
The x-axis of Figure~\ref{fig:diff-approx-plot} is the value of the \dispatchwidth parameter, and the y-axis is the predicted timing of \mca{} with that \dispatchwidth for the basic block consisting of the single instruction \lstinline{shrq $5, 16(
The blue points represent the prediction of \mca{} when instantiated with different values for \dispatchwidth.
The na\"{i}ve approach of optimizing \mca{} would be combinatorial search, since without a continuous and smooth surface to optimize, it is impossible to use standard first-order techniques.
\name{} instead first learns a \scaffold{} of \mca{}, represented by the orange line in Figure~\ref{fig:diff-approx-plot}.
This \scaffold{}, though not exactly the same as \mca{}, is smooth and differentiable.
Importantly, the \scaffold{} interpolates \mca{}'s predictions even in places where \mca{} does not have a defined output, such as between the integer-valued parameter settings.
Together, these properties mean that it is possible to optimize parameters against the \scaffold{} with first-order techniques like gradient descent.

\begin{figure}
  \includegraphics[width=\columnwidth]{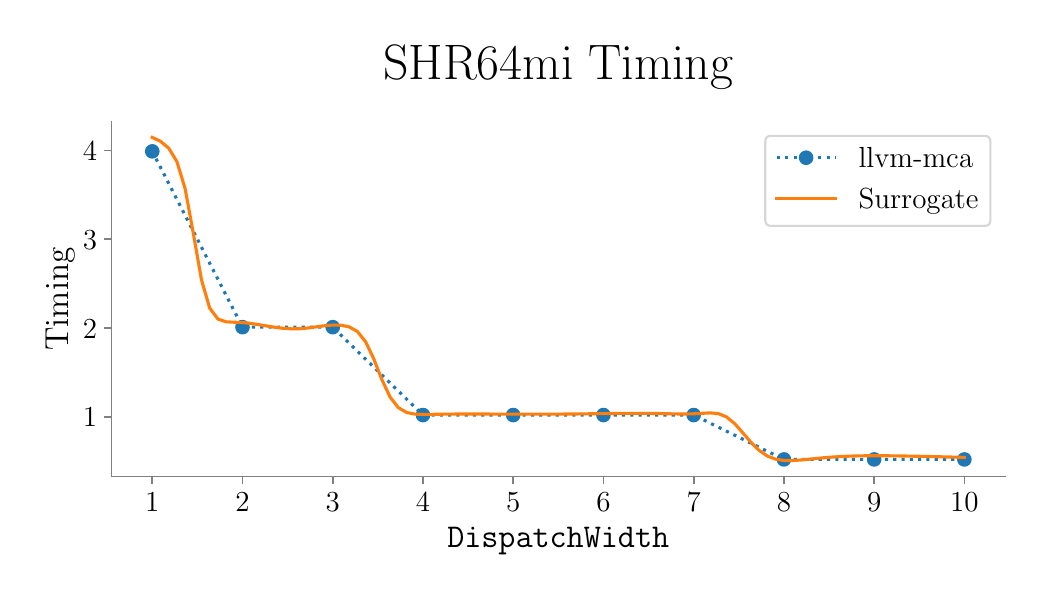}
  \caption{Example of timing predicted by \mca{} (blue) and a \scaffold (orange), while varying \dispatchwidth. By learning the \scaffold, we are able to optimize the parameter value with gradient descent, rather than requiring combinatorial search.}
  \label{fig:diff-approx-plot}
\end{figure}

\begin{figure*}
  \includegraphics[width=\textwidth]{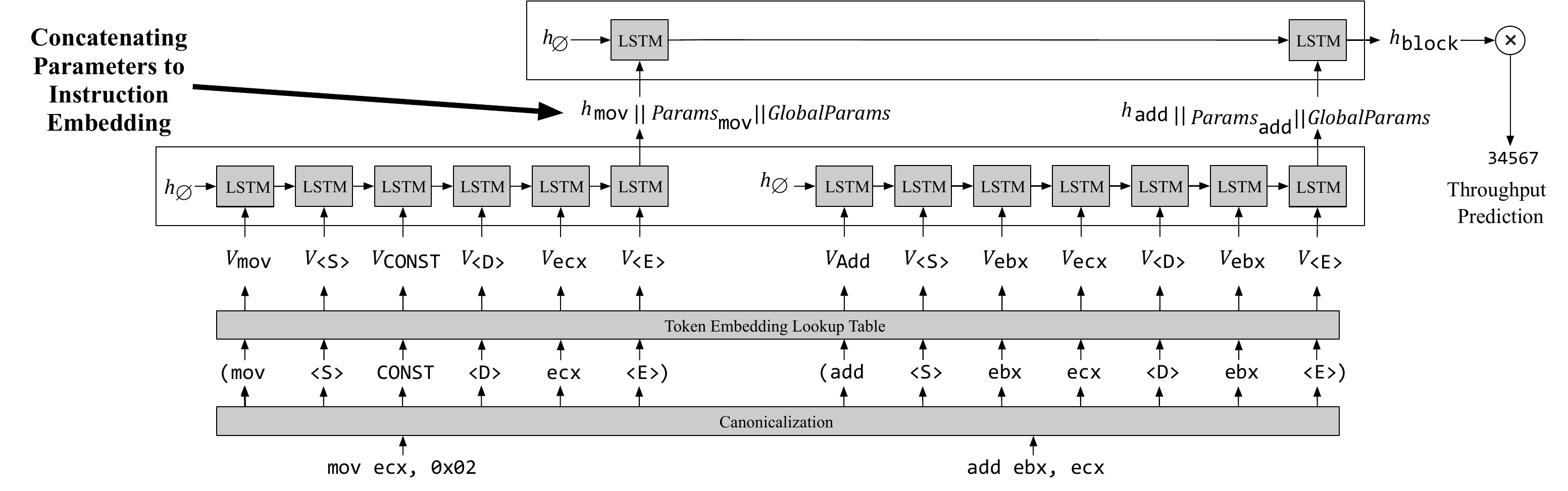}
  \caption{Design of the \scaffold, from \citet{mendis_ithemal_2019} with added parameter inputs. We use $\parallel$ to denote concatenation of parameters to the instruction embedding.}
  \label{fig:ithemal-parametric}
\end{figure*}

\section{Implementation}
\label{sec:implementation}

This section discusses our implementation of \name{}, available online at \url{https://github.com/ithemal/DiffTune}.

\paragraph{Parameters}
We consider two types of parameters that we optimize with \name{}: \emph{per-instruction parameters}, which are a uniform length vector of parameters associated with each individual instruction opcode (e.g.\ for \mca{}, a vector containing \latency, \nummicroops, etc.); and \emph{global parameters}, which are a vector of parameters that are associated with the overall simulator behavior (e.g.\ for \mca{}, a vector containing the \dispatchwidth and \reorderbuffersize).
We further support two types of constraints in our implementation: \emph{lower-bounded}, specifying that parameter values cannot be below a certain value (often $0$ or $1$), and \emph{integer-valued}, specifying that parameter values must be integers.
During optimization, all parameters are represented as floating-point.

\paragraph{\Scaffold design}
\Cref{fig:ithemal-parametric} presents our \scaffold design, which is capable of learning parameters for x86 basic block performance models such as \mca{}.

We use a modified version of Ithemal~\citep{mendis_ithemal_2019}, a learned basic block performance model, as the \scaffold.
In the standard implementation of Ithemal (without our modifications), Ithemal first uses an embedding lookup table to map the opcode and operands of each instruction into vectors.
Next, Ithemal processes the opcode and operand embeddings for each instruction with an LSTM, producing a vector representing each instruction.
Then, Ithemal processes the sequence of instruction vectors with another LSTM, producing a vector representing the basic block.
Finally, Ithemal uses a fully connected layer to turn the basic block vector into a single number representing Ithemal's prediction for the timing of that basic block.

We modify Ithemal in two ways to act as the \scaffold.
First, we replace each individual LSTM with a set of 4 stacked LSTMs, a common technique to increase representative capacity~\citep{hermans_training_2013}, to give Ithemal the capacity to represent the dependency of the prediction on the input parameters as well as on the input basic block.\footnote{A stack of 4 LSTMs resulted in the best validation error for the surrogate.}
Second, to provide the parameters as input we concatenate the per-instruction parameters and the global parameters to each instruction vector before processing the instruction vectors with the instruction-level LSTM.

\paragraph{Solving the optimization problems}
Training the \scaffold requires first defining sampling distributions for each parameter (e.g., a bounded uniform distribution on integers).
We then generate a large simulated dataset by repeatedly sampling a basic block from the ground-truth dataset, sampling a parameter table from the defined sampling distributions, instantiating the simulator with the parameter table, and generating a prediction for the basic block.
We train the \scaffold using SGD against this simulated dataset.
During \scaffold training, for parameters constrained to be lower-bounded we subtract the lower bound before passing them as input to the \scaffold.

To train the parameter table, we first initialize it to a random sample from the parameter sampling distribution.
We generate predictions using the parameter table plugged into the trained \scaffold, and train the parameter table by using SGD against the ground-truth dataset.
Importantly, when training the parameter table, the weights of the \scaffold are not updated.
During parameter table training, for parameters constrained to be lower-bounded we take the absolute value of the parameters before passing them as input to the \scaffold.

\paragraph{Parameter extraction}
Once we have trained the \scaffold and the parameter table using the optimization process described in \Cref{sec:approach}, we extract the parameters from the parameter table and use them in the original simulator.
For parameters with lower bounds, we take the absolute value of the parameter in the learned parameter table, then add the lower bound.
For integer parameters, we round the learned parameter to the nearest integer.
We do not use any special technique to handle unseen opcodes in the test set, just using the parameters for that opcode from the randomly initialized parameter table.

\begin{table*}
  \caption{Parameters learned for \mca{}.}
  \label{tab:mca-parameters}
  \centering
  \begin{tabularx}{\textwidth}{cccX}
    \toprule \textbf{Parameter} & \textbf{Count} & \textbf{Constraint} & \textbf{Description} \\ \midrule
    \dispatchwidth & 1 global & Integer, $\geq 1$ & How many micro-ops can be dispatched each cycle in the dispatch stage. \\
    \reorderbuffersize & 1 global & Integer, $\geq 1$ & How many micro-ops can fit in the reorder buffer. \\
    \nummicroops & 1 per-instruction & Integer, $\geq 1$ & How many micro-ops each instruction contains. \\
    \latency & 1 per-instruction & Integer, $\geq 0$ & The number of cycles before destination operands of that instruction can be read from. A latency value of 0 means that dependent instructions do not have to wait before being issued, and can be issued in the same cycle. \\
    \readadvance & 3 per-instruction & Integer, $\geq 0$ & How much to decrease the effective \latency of source operands. \\
    \portmap & 10 per-instruction & Integer, $\geq 0$ & The number of cycles the instruction occupies each execution port for. Represented as a 10-element vector per-instruction, where element $i$ is the number of cycles for which the instruction occupies port $i$.
    \\ \bottomrule
  \end{tabularx}
\end{table*}

\section{Evaluation}
\label{sec:results-main}
In this section, we report and analyze the results of using \name{} to learn the parameters of \mca{} across different x86 microarchitectures.
We first describe the methodological details of our evaluation in \Cref{sec:methodology}.
We then analyze the error of \mca{} instantiated with the learned parameters, finding the following:
\begin{itemize}
\item \name{} is able to learn parameters that lead to lower error than the default expert-tuned parameters across all four tested microarchitectures. (\Cref{sec:results-error})
\item Black-box global optimization with OpenTuner~\citep{ansel_opentuner_2014} cannot find a full set of parameters for \mca{}’s Intel x86 simulation model that match \mca{}'s default error. (\Cref{sec:results-opentuner})
\end{itemize}

To show that our implementation of \name{} is extensible to CPU simulators other than \mca{}, we evaluate \name{} on \exegesis{} in \Cref{sec:exegesis}.

\subsection{Methodology}
\label{sec:methodology}

Following \citet{chen_bhive_2019}, we use \mca{} version 8.0.1 (commit hash \texttt{19a71f6}).
We specifically focus on \mca{}'s Intel x86 simulation model:
\mca{} supports behavior beyond that described in \Cref{sec:background} (e.g., optimizing zero idioms, constraining the number of physical registers available, etc.) but this behavior is disabled by default in the Intel microarchitectures evaluated in this paper.
We do not enable or learn any behavior not present in \mca{}'s default Intel x86 simulation model, including when evaluating on AMD.

\paragraph{\mca{} parameters}
For each microarchitecture, we learn the parameters specified in \Cref{tab:mca-parameters}.
Following the default value in \mca{} for Haswell, we assume that there are 10 execution ports available for dispatch for all microarchitectures.
\mca{} supports simulation of instructions that can be dispatched to multiple different ports in the \portmap parameter.
However, the simulation of port group parameters in the \portmap does not correspond to standard definitions of port groups~\citep{andrea_email,agner,ritter_pmevo_2020}.
We therefore set all port group parameters in the \portmap to zero, removing that component of the simulation.

\paragraph{Dataset}
We use the BHive dataset from \citet{chen_bhive_2019}, which contains basic blocks sampled from a diverse set of applications (e.g., OpenBLAS, Redis, LLVM, etc.) along with the measured execution times of these basic blocks unrolled in a loop.
These measurements are designed to conform to the same modeling assumptions made by \mca{}.

\begin{table}
  \caption{Dataset summary statistics.}
  \label{tab:dataset-summary-statistics}
  \centering
  \begin{tabular}{lc}
    \toprule
    \textbf{Statistic} & \textbf{Value} \\
    \midrule
    \# Blocks &  \\
    \multicolumn{1}{r}{Train} & $230111$ \\
    \multicolumn{1}{r}{Validation} & $28764$ \\
    \multicolumn{1}{r}{Test} & $28764$ \\
    \multicolumn{1}{r}{\textbf{Total}} & \textbf{$287639$} \\ \midrule
    Block Length & \\
    \multicolumn{1}{r}{Min} & $1$ \\
    \multicolumn{1}{r}{Median} & $3$ \\
    \multicolumn{1}{r}{Mean} & $4.93$ \\
    \multicolumn{1}{r}{Max} & $256$ \\ \midrule
    Median Block Timing & \\
    \multicolumn{1}{r}{Ivy Bridge} & $132$ \\
    \multicolumn{1}{r}{Haswell} & $123$ \\
    \multicolumn{1}{r}{Skylake} & $120$ \\
    \multicolumn{1}{r}{Zen 2} & $114$ \\ \midrule
    \# Unique Opcodes & \\
    \multicolumn{1}{r}{Train} & $814$ \\
    \multicolumn{1}{r}{Val} & $610$ \\
    \multicolumn{1}{r}{Test} & $580$ \\
    \multicolumn{1}{r}{\textbf{Total}} & $837$ \\
    \bottomrule
  \end{tabular}
\end{table}

We use the latest available version of the released timings on Github.\footnote{\url{https://github.com/ithemal/bhive/tree/5878a18/benchmark/throughput}}
We evaluate on the datasets released with BHive for the Intel x86 microarchitectures Ivy Bridge, Haswell, and Skylake.
We also evaluate on AMD Zen~2, which was not included in the BHive dataset.
The Zen~2 measurements were gathered by running a version of BHive modified to time basic blocks using AMD performance counters on an AMD EPYC 7402P, using the same methodology as \citeauthor{chen_bhive_2019}.
Following \citeauthor{chen_bhive_2019}, we remove all basic blocks potentially affected by virtual page aliasing.

We randomly split off $80\%$ of the measurements into a training set, $10\%$ into a validation set for development, and $10\%$ into the test set reported in this paper.
We use the same train, validation, and test set split for all microarchitectures.
The training and test sets are block-wise disjoint: there are no identical basic blocks between the training and test set.
Summary statistics of the dataset are presented in \Cref{tab:dataset-summary-statistics}.

\paragraph{Objective}

We use the same definition of timing as \citet{chen_bhive_2019}: the number of cycles it takes to execute 100 iterations of the given basic block, divided by 100.
Following \citeauthor{chen_bhive_2019}'s definition of error, we optimize \mca{} to minimize the mean absolute percentage error (MAPE) against a dataset: \[
  \text{Error} \triangleq \frac{1}{|\mathcal{D}|}\sum_{(x, y) \in \mathcal{D}} \frac{|f(x) - y|}{y}
\]
We note that an error of above $100\%$ is possible when $f(x)$ is much larger than $y$.

\paragraph{Training methodolgy}
We use Pytorch-1.2.0 on an NVIDIA Tesla V100 to train the \scaffold and parameters.

We train the \scaffold and the parameter table using Adam~\citep{kingma_adam_2014}, a stochastic first-order optimization technique, with a batch size of 256.
We use a learning rate of $0.001$ to train the \scaffold and a learning rate of $0.05$ to train the parameter~table.

To train the \scaffold, we generate a simulated dataset of $2301110$ blocks ($10\times$ the length of the original training set).
For each basic block in the simulated dataset, we sample a random parameter table, with each \latency a uniformly random integer between $0$ and $5$ (inclusive), each value in the \portmap uniform between $0$ and $2$ cycles to between $0$ and $2$ randomly selected ports for each instruction, each \readadvance between $0$ and $5$, each \nummicroops between $1$ and $10$, the \dispatchwidth uniform between $1$ and $10$, and the \reorderbuffersize uniform between $50$~and~$250$.
A random parameter table sampled from this distribution has error $171.4\% \pm 95.7\%$.
See \Cref{sec:limitations} for more discussion of these sampling distributions.

We loop over this simulated dataset $6$ times when training the \scaffold, totaling an equivalent of $60$ epochs over the original training set.
To train the parameter table, we initialize it to a random sample from the parameter training distribution, then train it for $1$ epoch against the original training set.

\subsection{Error of Learned Parameters}
\label{sec:results-error}

\begin{table}
\caption{Error of \mca{} with the default and learned parameters, compared against baselines.}
\label{tab:results-mca-all}
  \centering
  \begin{tabular}{p{12em}cc}
  \toprule
  \textbf{Architecture} \hfill \textbf{Predictor} & \textbf{Error} & \textbf{Kendall's Tau} \\
    \midrule \textbf{Ivy Bridge}
    \hfill Default & $33.5\%$ & 0.788 \\
    \hfill \name{} & $25.4\% \pm 0.5\%$  & $0.735 \pm 0.012$ \\ \cmidrule{2-3}
    \hfill Ithemal & $9.4\%$ & 0.858 \\
    \hfill IACA & $15.7\%$ & $0.810$ \\
    \hfill OpenTuner & $102.0\%$ & 0.515 \\
    \midrule    \textbf{Haswell}
    \hfill Default & $25.0\%$ & 0.783 \\
    \hfill \name{} & $23.7\% \pm 1.5\%$ & $0.745 \pm 0.009$ \\ \cmidrule{2-3}
    \hfill Ithemal & $9.2\%$ & 0.854 \\
    \hfill IACA & $17.1\%$ & $0.800$ \\
    \hfill OpenTuner & $105.4\%$ & 0.522 \\
    \midrule    \textbf{Skylake}
    \hfill Default & $26.7\%$ & 0.776 \\
    \hfill \name{} & $23.0\% \pm 1.4\%$ & $0.748 \pm 0.008$ \\ \cmidrule{2-3}
    \hfill Ithemal & $9.3\%$ & 0.859 \\
    \hfill IACA & $14.3\%$ & $0.811$ \\
    \hfill OpenTuner & $113.0\%$ & 0.516 \\
    \midrule    \textbf{Zen 2}
    \hfill Default & $34.9\%$\footnotemark & 0.794 \\
    \hfill \name{} & $26.1\% \pm 1.0\%$ & $0.689 \pm 0.007$ \\ \cmidrule{2-3}
    \hfill Ithemal & $9.4\%$ & 0.873 \\
    \hfill IACA & N/A & N/A \\
    \hfill OpenTuner & $131.3\%$ & 0.494 \\
  \bottomrule
\end{tabular}%
\end{table}

\Cref{tab:results-mca-all} presents the average error and correlation of \mca{} with the default parameters (labeled default), \mca{} with the learned parameters (labeled \name{}).
As baselines, \Cref{tab:results-mca-all} also presents
Ithemal's error, as the most accurate predictor evaluated by \citeauthor{chen_bhive_2019},
IACA's error, as the most accurate analytical model evaluated by \citeauthor{chen_bhive_2019},
and \mca{} with parameters learned by OpenTuner (which we discuss further in \Cref{sec:results-opentuner}).
Because IACA is written by Intel to analyze Intel microarchitectures, it does not generate predictions for Zen~2.
We report mean absolute percentage error, as defined in \Cref{sec:methodology}, and Kendall's Tau rank correlation coefficient, measuring the fraction of pairs of timing predictions in the test set that are ordered correctly.
For the learned parameters, we report the mean and standard deviation of error and Kendall's Tau across three independent runs~of~\name{}.

Across all microarchitectures, the parameter set learned by \name{} achieves equivalent or better error than the default parameter set.
These results demonstrate that \name{} can learn all of \mca{}’s microarchitecture-specific parameters, from scratch, to equivalent accuracy as the hand-written parameters.

\begin{table}
  \caption{Error of \mca{} with default and learned parameters on Haswell, grouped by BHive applications~and~categories.}
  \label{tab:bhive-category-errors}

  \centering
  \begin{tabular}{lccc}
    \toprule
    \multirow{2}{*}{\textbf{Block Type}} & \multirow{2}{*}{\textbf{\# Blocks}} & \textbf{Default} &  \textbf{Learned} \\
    &&\textbf{Error}&\textbf{Error}\\
\midrule
    OpenBLAS & 1478 & $28.8\%$ & $29.0\%$ \\
Redis & 839 & $41.2\%$ & $22.5\%$ \\
SQLite & 764 & $32.8\%$ & $21.6\%$ \\
GZip & 182 & $40.6\%$ & $20.6\%$ \\
TensorFlow & 6399 & $33.5\%$ & $22.1\%$ \\
Clang/LLVM & 18781 & $22.0\%$ & $21.0\%$ \\
Eigen & 387 & $44.3\%$ & $23.8\%$ \\
Embree & 1067 & $34.1\%$ & $21.3\%$ \\
FFmpeg & 1516 & $30.9\%$ & $21.2\%$ \\
\midrule
Scalar (Scalar ALU operations)     &   7952 &            $17.2\%$ &            $18.9\%$ \\
Vec (Purely vector instructions)        &    104 &            $35.3\%$ &            $39.6\%$ \\
    Scalar/Vec  &    \multirow{2}{*}{614} &            \multirow{2}{*}{$53.6\%$} &            \multirow{2}{*}{$37.5\%$} \\
\hfill    (Scalar and vector arithmetic) \\
Ld (Mostly loads)         &  10850 &            $27.2\%$ &            $24.4\%$ \\
St (Mostly stores)        &   4490 &            $24.7\%$ &            $08.7\%$ \\
Ld/St (Mix of loads and stores)      &   4754 &            $27.9\%$ &            $30.3\%$ \\
\bottomrule
  \end{tabular}
\end{table}

\footnotetext{llvm-8.0.1 does not support Zen~2. This default error we report for Zen~2 uses the znver1 target in llvm-8.0.1, targeting Zen~1. The Zen~2 target in llvm-10.0.1 has a higher error of $39.8\%$.}

We also analyze the error of \mca{} on the Haswell microarchitecture using the evaluation metrics from \citet{chen_bhive_2019}, designed to validate x86 basic block performance models.
\citeauthor{chen_bhive_2019} present three forms of error analysis: overall error, per-application error, and per-category error.
Overall error is the error reported in \Cref{tab:results-mca-all}.
Per-application error is the average error of basic blocks grouped by the source application of the basic block (e.g., TensorFlow, SQLite, etc.; blocks can have multiple different source applications).
Per-category error is the average error of basic blocks grouped into clusters based on the hardware resources used by each basic block.

The per-application and per-category errors are presented in \Cref{tab:bhive-category-errors}.
The learned parameters outperform the defaults across most source applications, with the exception of OpenBLAS where the learned parameters result in $0.2\%$ higher error.
The learned parameters perform similarly to the default across most categories, with the primary exceptions of the Scalar/Vec category and the St category, in which the learned parameters perform significantly better than the default parameters.

\subsection{Black-box global optimization with OpenTuner}
\label{sec:results-opentuner}
In this section, we describe the methodology and performance of using black-box global optimization with OpenTuner~\citep{ansel_opentuner_2014} to find parameters for \mca{}.
We find that OpenTuner is not able to find parameters that lead to equivalent error as \name{} in \mca{}'s Intel x86 simulation~model.

\paragraph{Background}
We use OpenTuner as a representative example of a black-box global optimization technique.
OpenTuner is primarily a system for tuning parameters of programs to decrease run-time (e.g., tuning compiler flags, etc.), but has also been validated on other optimization problems, such as finding the series of button presses in a video game simulator that makes the most progress in the game.

OpenTuner is an iterative algorithm that uses a multi-armed bandit to pick the most promising search technique among an ensemble of search techniques that span both convex and non-convex optimization.
On each iteration, the bandit evaluates the current set of parameters.
Using the results of that evaluation, the bandit then selects a search technique that then proposes a new set of parameters.

\paragraph{Methodology}
For computational budget parity with \name{}, we permit OpenTuner to evaluate the same number of basic blocks as used end-to-end in our learning approach.
We initialize OpenTuner with a sample from \name{}'s parameter table sampling distribution.
We constrain OpenTuner to search per-instruction (\nummicroops, \latency, \readadvance, \portmap) parameter values between 0 and 5, \dispatchwidth between 1 and 10, and \reorderbuffersize between 50 and 250; these ranges contain the majority of the parameter values observed in the default and learned parameter sets.\footnote{Widening the search space beyond this range resulted in a significantly higher error for OpenTuner.}
We evaluate the accuracy of \mca{} with the resulting set of parameters using the same methodology as in \Cref{sec:results-error}.

\paragraph{Results}
\Cref{tab:results-mca-all} presents the error of \mca{} when parameterized with OpenTuner's discovered parameters.
OpenTuner's parameters perform worse than those of \name{}, resulting in error above $100\%$ across all microarchitectures.
Thus, \name{} requires substantially fewer examples to optimize \mca{} than OpenTuner requires.

\begin{figure*}
  \begin{center}
    \centering
    \begin{subfigure}[b]{\columnwidth}
      \includegraphics[width=\textwidth]{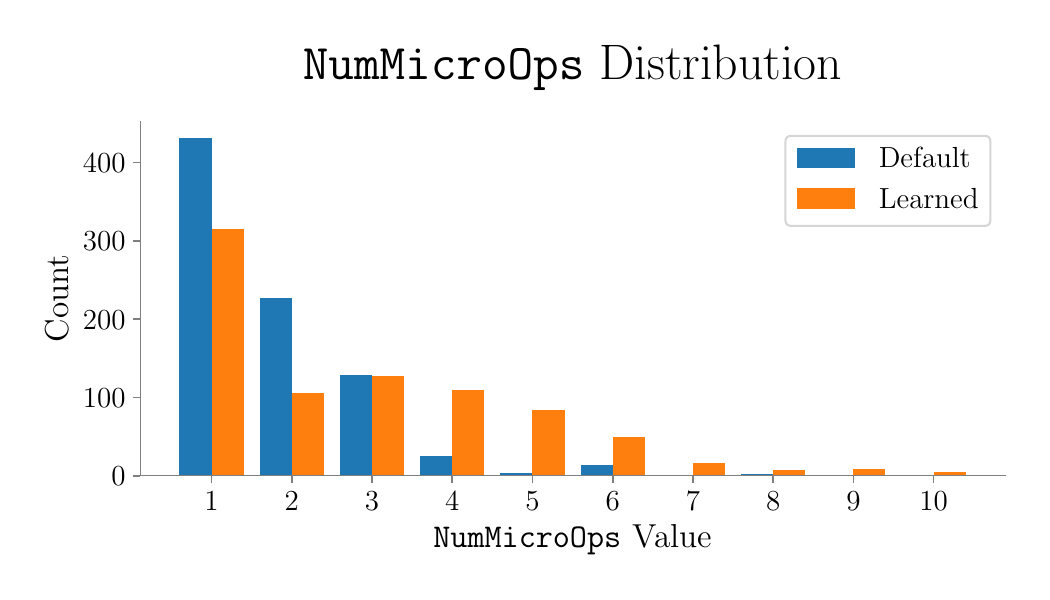}
      \caption{Distribution of default and learned \nummicroops values.}
      \label{fig:nummicroops-distr}
    \end{subfigure}
    \hfill
    \begin{subfigure}[b]{\columnwidth}
      \includegraphics[width=\textwidth]{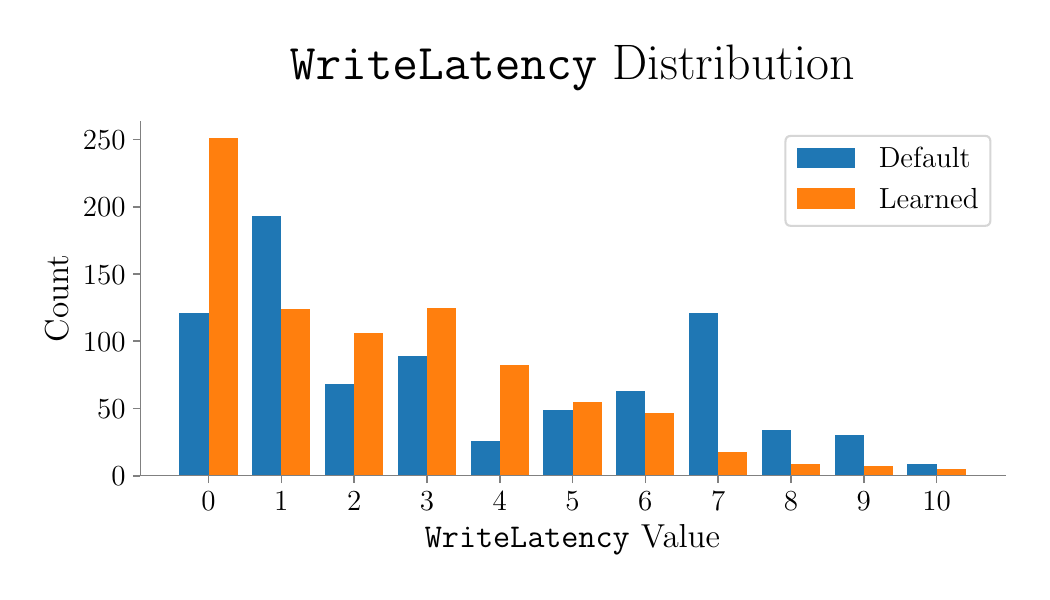}
      \caption{Distribution of default and learned \latency values.}
      \label{fig:latency-distr}
    \end{subfigure}
    \\
    \begin{subfigure}[b]{\columnwidth}
      \includegraphics[width=\textwidth]{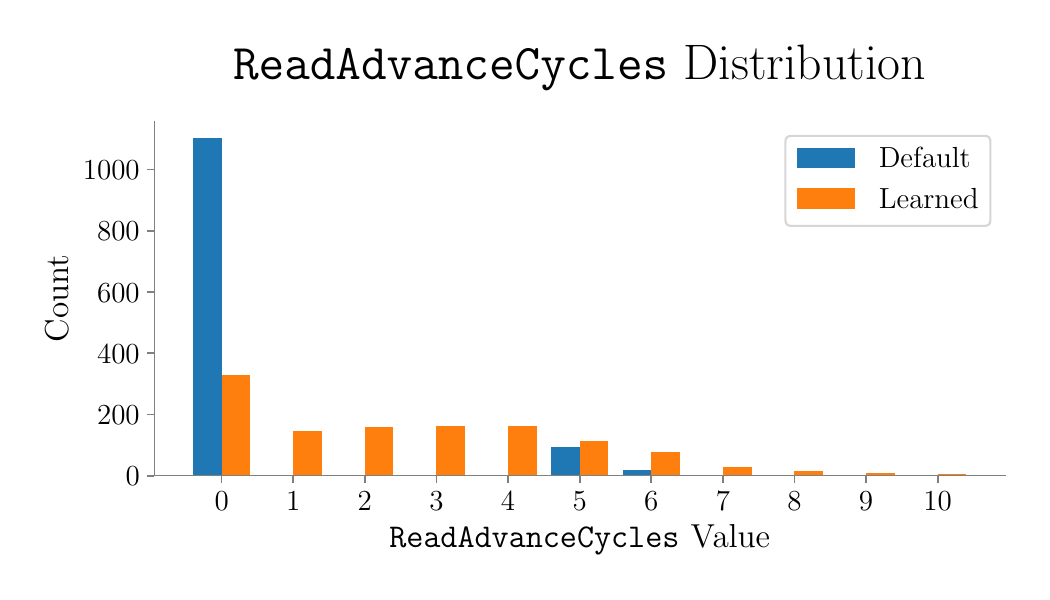}
      \caption{Distribution of default and learned \readadvance values.}
      \label{fig:readadvance-distr}
    \end{subfigure}
    \hfill
    \begin{subfigure}[b]{\columnwidth}
      \includegraphics[width=\textwidth]{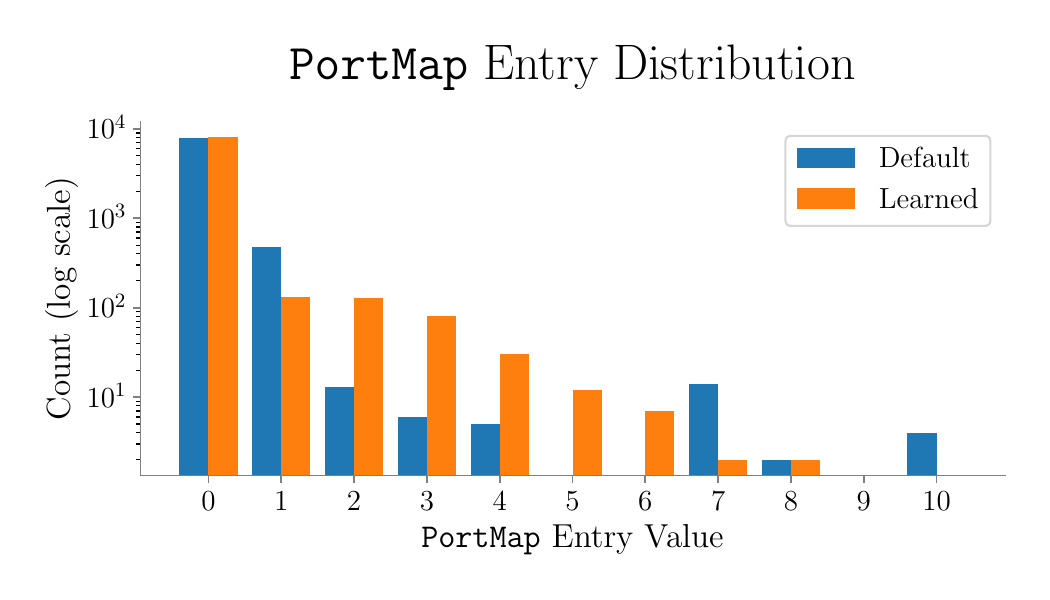}
      \caption{Distribution of default and learned \portmap values.}
      \label{fig:portmap-distr}
    \end{subfigure}
  \end{center}
  \caption{Distributions of default and learned parameter values on Haswell.}
  \label{fig:distributions}
\end{figure*}

\section{Analysis}
\label{sec:analysis}
In this section, we analyze the parameters learned by \name{} on \mca{}, answering the following research questions:
\begin{itemize}
\item How similar are the learned parameters to the default parameters in \mca{}? (\Cref{sec:results-default-comparison})
\item How optimal are the learned parameters? (\Cref{sec:results-optimality})
\item How semantically meaningful are the learned parameters? (\Cref{sec:results-semantics})
\end{itemize}

\subsection{Comparison of Learned Parameters to Defaults}
\label{sec:results-default-comparison}

This section compares the default parameters to the learned parameters (from a single run of \name{}) in Haswell.

\paragraph{Distributional similarities}
To determine the distributional similarity of the learned parameters to the default parameters, \Cref{fig:distributions} shows histograms of the values of the default and learned per-instruction parameters (\nummicroops, \latency, \readadvance, \portmap).
The primary distinctions between the distributions are in \latency and \readadvance; the learned parameters otherwise follow similar distributions to the defaults.

The distributions of default and learned \latency values in \Cref{fig:latency-distr} primarily differ in that only 1 out of the 837 opcodes in the default Haswell parameters has \latency 0 (\texttt{VZEROUPPER}), whereas 251 out of the 837 opcodes in the learned parameters have \latency 0.
As discussed in \Cref{tab:mca-parameters}, a \latency value of 0 means that dependent instructions do not have to wait before being issued, and can be issued in the same cycle; instructions may still be bottlenecked elsewhere in the simulation pipeline (e.g., in the execute stage).

The distributions of default and learned \readadvance are presented in \Cref{fig:readadvance-distr}.
The default \readadvance are mostly 0, with a small population having values 5 and 7; in contrast, the learned \readadvance are fairly evenly distributed, with a plurality being 0.
Given that a significant fraction of learned \latency values are 0, it is likely that many \readadvance values have little to no effect.\footnote{As noted in \Cref{sec:background}, \mca{} subtracts \readadvance from \latency to compute a dependency chain's latency. The result of this subtraction is clipped to be no less than zero.}

\begin{table}
  \caption{Default and learned global parameters.}
  \label{tab:global}
  \begin{tabular}{p{12em}cc}
    \toprule
    \textbf{Architecture} \hfill \textbf{Parameters} & \textbf{\dispatchwidth} & \textbf{\reorderbuffersize} \\

    \midrule \textbf{Haswell}
    \hfill Default & 4 & 192 \\
    \hfill Learned & 4 & 144 \\

    \bottomrule
  \end{tabular}
\end{table}

\begin{figure}
  \centering
  \includegraphics[width=\columnwidth]{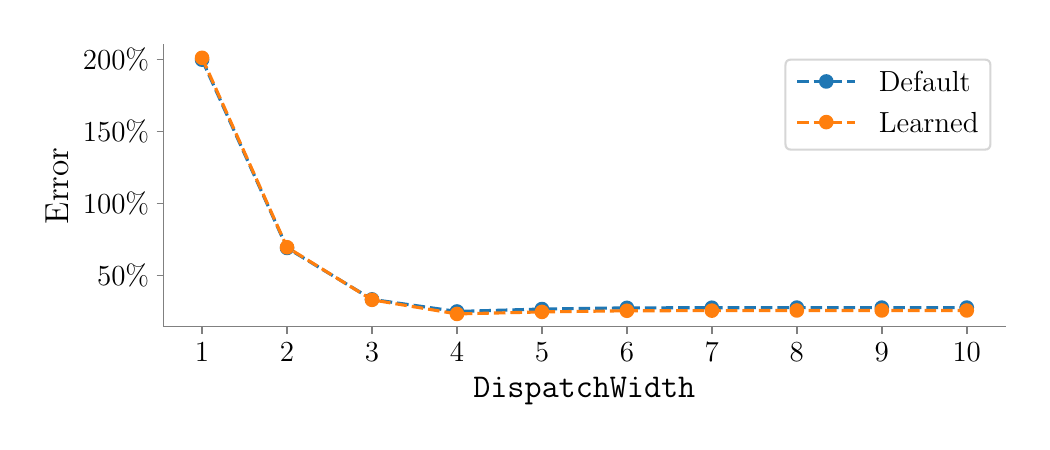} \\
  \includegraphics[width=\columnwidth]{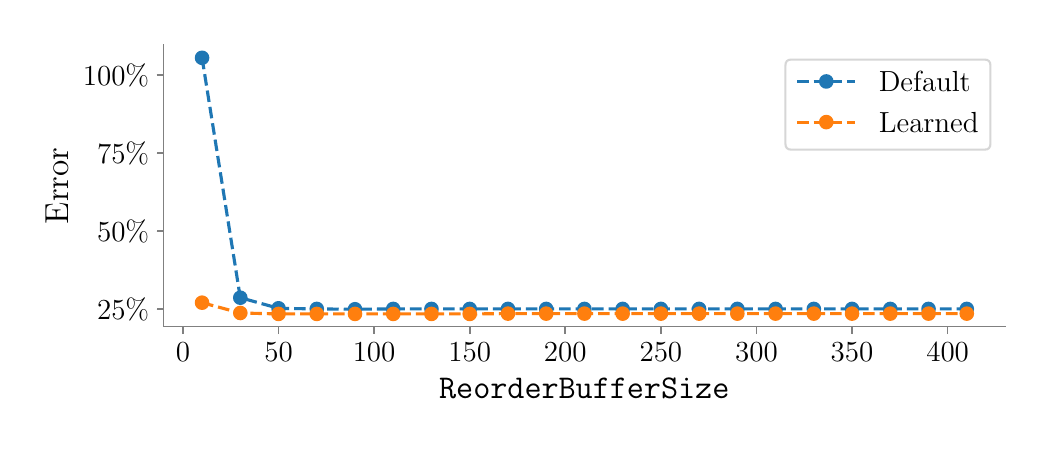}
	\caption{\mca{}'s sensitivity to values of \dispatchwidth (Top) and \reorderbuffersize (Bottom) within the default (Blue) and learned (Orange) parameters.}
  \label{fig:global-sensitivity}
\end{figure}

\paragraph{Global parameters}
\Cref{tab:global} shows the default and learned global parameters (\dispatchwidth and \reorderbuffersize).
The learned \dispatchwidth parameter is close to the default \dispatchwidth parameter, while the learned \reorderbuffersize parameter differs significantly from the default.
By analyzing \mca{}'s sensitivity to values of \dispatchwidth and \reorderbuffersize within the default and learned parameters in \Cref{fig:global-sensitivity}, we find that although the learned global parameters do not match the default values exactly, they approximately minimize \mca{}'s error because there is a wide range of values that result in approximately the same error.
While \mca{} is sensitive to small perturbations in the value of the \dispatchwidth parameter (with the default parameters, a \dispatchwidth of $3$ has error $33.5\%$, $4$ has error $25.0\%$, and $5$ has error $26.8\%$), it is relatively insensitive to perturbations of the \reorderbuffersize (with the default parameters, all \reorderbuffersize values above $70$ have error $25.0\%$).
This is primarily because one of \mca{}'s core modeling assumptions, that memory accesses always resolve in the L1 cache, means that most instructions spend few cycles in the issue, execute, and retire phases; the \reorderbuffersize is therefore rarely a bottleneck in \mca{}'s modeling of the~BHive~dataset.

\subsection{Optimality}
\label{sec:results-optimality}

This \namecref{sec:results-optimality} shows that while the parameters learned by \name{} match the error of the default parameters, the learned values are not optimal:
by using \name{} to optimize just a subset of \mca{}'s parameters, and keeping the rest as their expert-tuned default values, we are able to find parameters with lower error than when learning the entire set of parameters.

\paragraph{Experiment}
We learn only each instruction's \latency in \mca{}.
We keep all other parameters as their default values.
The dataset and objective used in this task are otherwise the same as presented in \Cref{sec:methodology}.

\paragraph{Methodology}
Training hyperparameters are similar to those presented in \Cref{sec:methodology}, and are reiterated here with modifications made to learn just \latency parameters.
We train both the \scaffold and the parameter table using Adam~\citep{kingma_adam_2014} with a batch size of 256.
We use a learning rate of $0.001$ to train the \scaffold and a learning rate of $0.05$ to train the parameter table.
To train the \scaffold, we generate a simulated dataset of $2301110$ blocks.
For each basic block in the simulated dataset, we sample a random parameter table, with each \latency a uniformly random integer between $0$ and $10$ (inclusive).
We loop over this simulated dataset $3$ times when training the \scaffold.
To train the parameter table, we initialize it to a random sample from the parameter training distribution, then train it for $1$ epoch against the original training set.

\paragraph{Results}
On Haswell, this application of \name{} results in an error of $16.2\%$ and a Kendall Tau correlation coefficient of $0.823$, compared to an error of $23.7\%$ and a correlation of $0.745$ when learning the full set of parameters with \name{}.
These results demonstrate that \name{} does not find a globally optimal parameter set when learning \mca{}'s full set of parameters.
This suboptimality is due in part to the non-convex nature of the problem and the size of the~parameter~space.

\subsection{Case Studies}
\label{sec:results-semantics}
This section presents case studies of basic blocks simulated with the default and with the learned parameters, showing where the learned parameters better reflect the ground truth data, and where the learned parameters reflect degenerate cases of the optimization algorithm.
To simplify exposition, the results in this section use just the learned \latency values from the experiment in \Cref{sec:results-optimality}.

\paragraph{\texttt{PUSH64r}}
The default \latency with the Haswell parameters for the \texttt{PUSH64r} opcode (push a 64-bit register onto the stack, decrementing the stack pointer) is 2 cycles.
In contrast, the \latency learned by \name{} is 0 cycles.
This leads to significantly more accurate predictions for blocks that contain \texttt{PUSH64r} opcodes, such as the following (in which the default and learned latency for \instr{testl} are both 1 cycle):
\begin{lstlisting}
        pushq	%rbx
        testl	%r8d, %r8d
\end{lstlisting}
The true timing of this block as measured by \citet{chen_bhive_2019} is 1.01 cycles.
On this block, \mca{} with the default Haswell parameters predicts a timing of 2.03 cycles: The \texttt{PUSH64r} forms a dependency chain with itself, so the default \latency before each \texttt{PUSH64r} can be dispatched is 2 cycles.
In contrast, \mca{} with the learned Haswell values predicts that the timing is 1.03 cycles, because the learned \latency is 0 meaning that there is no delay before the following \texttt{PUSH64r} can be issued, but the \portmap for \texttt{PUSH64r} still occupies \texttt{HWPort4} for a cycle before the instruction is retired; this 1-cycle dependency chain results in a more accurate prediction.
In this case, \name{} learns a \latency that leads to better accuracy for the \texttt{PUSH64r}~opcode.

\paragraph{\texttt{XOR32rr}}
The default \latency in Haswell for the \texttt{XOR32rr} opcode (xor two registers with each other) is 1 cycle.
The \latency learned by \name{} is again 0 cycles.
This is not always correct -- however, a common use of \texttt{XOR32rr} is as a \emph{zero idiom}, an instruction that sets a register to zero.
For example, \lstinline{xorq 
Intel processors have a fast path for zero idioms -- rather than actually computing the \instr{xor}, they simply set the value to zero.
Most of the instances of \texttt{XOR32rr} in our dataset ($4047$ out of $4218$) are zero idioms.
This leads to more accurate predictions in the general case, as can be seen in the following example:
\begin{lstlisting}
        xorl	%r13d, %r13d
\end{lstlisting}
The true timing of this block is 0.31 cycles.
With the default \latency value of 1, the Intel x86 simulation model of \mca{} does not recognize this as a zero idiom and predicts that this block has a timing of 1.03 cycles.
With the learned \latency value of 0 and the fact that there are no bottlenecks specified by the \portmap, \mca{} executes the \instr{xor}s as quickly as possible, bottlenecked only by the \nummicroops of $1$ and the \dispatchwidth of 4.
With this change, \mca{} predicts that this block has a timing of 0.27 cycles, again closer to the ground truth.

\paragraph{\texttt{ADD32mr}}
Unfortunately, it is impossible to distinguish between semantically meaningful values that make the simulator more correct, and degenerate values that improve the accuracy of the simulator without adding interpretability.
For instance, consider \texttt{ADD32mr}, which adds a register to a value in memory and writes the result back to memory:
\begin{lstlisting}
        addl	%eax, 16(%rsp)
\end{lstlisting}
This block has a true timing of 5.97 cycles because it is essentially a chained load, add, then store---with the L1 cache latency being 4 cycles.
However, \mca{} does not recognize the dependency chain this instruction forms with itself, so even with the default Haswell \latency of 7 cycles for \texttt{ADD32mr}, \mca{} predicts that this block has an overall timing of 1.09 cycles.
Our methodology recognizes the need to predict a higher timing, but is fundamentally unable to change a parameter in \mca{} to enable \mca{} to recognize the dependency chain (because no such parameter exists).
Instead, our methodology learns a degenerately high \latency of 62 for this instruction, allowing \mca{} to predict an overall timing of 1.64 cycles, closer to the true value.
This degenerate value increases the accuracy of \mca{} without leading to semantically useful \latency parameters.
This case study shows that the interpretability of the learned parameters is only as good as the simulation fidelity; when the simulation is a poor approximation to the physical behavior of the CPU, the learned parameters do not correspond to semantically~meaningful~values.

\section{Future Work}
\label{sec:limitations}

\name{} is an effective technique to learn simulator parameters, as we demonstrate with \mca{} (\Cref{sec:results-main}) and \exegesis{} (\Cref{sec:exegesis}).
However, there are several aspects of \name{}'s approach that are designed around the fact that \mca{} and \exegesis{} are basic block simulators that are primarily parameterized by ordinal parameters with few constraints between the values of individual parameters.
We believe that \name{}'s overall approach---differentiable surrogates---can be extended to whole program and full system simulators that have richer parameter spaces, such as gem5, by extending a subset of \name{}'s individual~components.

\paragraph{Whole program and full system simulation}
\name{} requires a differentiable surrogate that can learn the simulator's behavior to high accuracy.
Ithemal~\citep{mendis_ithemal_2019}---the model we use for the {\scaffold}---operates on basic blocks with the assumption that all data accesses resolve in the L1 cache, which is compatible with our evaluation of \mca{} and \exegesis{} (which make the same assumptions).
While Ithemal could potentially model whole programs (e.g., branching) and full systems (e.g., cache behavior) with limited modifications, it may require significant extensions to learn such more~complex~behavior~\citep{vila_cachequery_2020, hashemi_learning_2018}.

In addition to the design of the \scaffold, training the \scaffold would require a new dataset that includes whole programs, along with any other behavior modeled by the simulator being optimized (e.g., memory).
Acquiring such a dataset would require extending timing methodologies like BHive~\citep{chen_bhive_2019} to the full scope of target behavior.

\paragraph{Non-ordinal parameters}
\name{} only supports ordinal parameters and does not support categorical or boolean parameters.
\name{} requires a relaxation of discrete parameters to continuous values to perform optimization, along with a method to extract the learned relaxation back into the discrete parameter type (e.g., \name{} relaxes integers to real numbers, and extracts the learned parameters by rounding back to integers).
Supporting categorical and boolean parameters would require designing and evaluating a scheme to represent and extract such parameters within \name{}.
One candidate representation is one-hot encoding,
but has not been evaluated in \name{}.

\paragraph{Dependent parameters}
All integers in the range $[1, \infty)$ are valid settings for \mca{}'s parameters.
However, other simulators, such as gem5, have stricter conditions---expressed as assertions in the simulator---on the relationship among different parameters.\footnote{For an example, see \url{https://github.com/gem5/gem5/blob/v20.0.0.0/src/cpu/o3/decode_impl.hh\#L423}, which is based on the interaction between different parameters, defined at \url{https://github.com/gem5/gem5/blob/v20.0.0.0/src/cpu/o3/decode_impl.hh\#L75}.}
\name{} also does not apply when there is a variable number of parameters: we are able to learn the port mappings in a fixed-size \portmap, but do not learn the number of ports in the \portmap, instead fixing it at 10 (the default value for the Haswell microarchitecture).
Extending \name{} to optimize simulators with rich, dynamic constrained relationships between parameters motivates new work in efficient techniques to sample such sets of parameters~\citep{dutra_efficient_2018}.

\paragraph{Sampling distributions}
Extending \name{} to other simulators also requires defining appropriate sampling distributions for each parameter.
While the sampling distributions do not have to directly lead to parameter settings that lead the simulator to have low error (e.g., the sampling distributions defined in \Cref{sec:methodology} lead to an average error of \mca{} on Haswell of $171.4\% \pm 95.7\%$), they do need to contain values that the parameter table estimate may take on during the parameter table optimization phase (because neural networks like our modification of Ithemal are not guaranteed to be able to accurately extrapolate outside of their training distribution).
Other approaches to optimizing with learned differentiable surrogates handle this by continuously re-optimizing the \scaffold{} in a region around the current parameter estimate~\citep{shirobokov_differentiating_2020}, a promising direction that could alleviate the need to hand-specify proper sampling distributions.

\section{Related Work}
\label{sec:related}

Simulators are widely used for architecture research to model the interactions of architectural components of a system~\citep{binkert_gem5_2011,llvm-mca,ptlsim,marss,sanchez_zsim_2013}.
Configuring and validating CPU simulators to accurately model systems of interest is a challenging task~\citep{chen_bhive_2019,gutierrez_sources_2014,akram_validation_2019}.
We review related techniques for setting CPU simulator parameters in \Cref{sec:related-simulate}, as well as related techniques to \name{}~in~\Cref{sec:related-optimize}.

\subsection{Setting CPU Simulator Parameters}
\label{sec:related-simulate}
In this section, we discuss related approaches for setting CPU simulator parameters.

\paragraph{Measurement}
One methodology for setting the parameters of an abstract model is to gather fine-grained measurements of each individual parameter's realization in the physical machine~\citep{agner,abel_uops_2019} and then set the parameters to their measured values~\citep{mca-agner,mca-uops}.
When the semantics of the simulator and the semantics of the measurement methodology coincide, then these measurements can serve as effective parameter values.
However, if there is a mismatch between the simulator and measurement methodology, then measurements may not provide effective parameter settings.

All fine-grained measurement frameworks rely on accurate hardware performance counters to measure the parameters of interest.
Such performance counters do not always exist, such as with per-port measurement performance counters on AMD Zen~\citep{ritter_pmevo_2020}.
When such performance counters are present, they are not always reliable~\citep{weaver_hardware_2008}.

\paragraph{Optimizing CPU simulators}
Another methodology for setting parameters of an abstract model is to infer the parameters from end-to-end measurements of the performance of the physical machine.
In the most related effort in this space, \citet{ritter_pmevo_2020} present a framework for inferring port usage of instructions based on optimizing against a CPU model that solves a linear program to predict the throughput of a basic block. Their approach is specifically designed to infer port mappings and it is not clear how the approach could be extended to infer different parameters in a more complex simulator, such as extending their simulation model to include data dependencies, dispatch width, or reorder buffer size.
To the best of our knowledge, \name{} is the first approach designed to optimize an arbitrary simulator, provided that the simulator and its parameters match \name{}'s scope of applicability~(\Cref{sec:limitations}).

\subsection{Differentiable surrogates and approximations}
\label{sec:related-optimize}

In this section, we survey techniques related to \name{} that facilitate optimization by using differentiable surrogates or approximations.

\paragraph{Optimization with learned differentiable surrogates}
Optimization of black-box and non-differentiable functions with learned differentiable surrogates is an emerging set of techniques, with applications in computer graphics~\citep{tseng_hyperparameter_2019}, physical sciences~\citep{shirobokov_differentiating_2020,louppe_adversarial_2017}, reinforcement learning~\citep{grathwohl_backpropagation_2018}, and computer security~\citep{she_neuzz_2019}.
\citet{tseng_hyperparameter_2019} use a convolutional neural network as a learned differentiable surrogate to optimize 32 parameters in a black-box camera imaging pipeline.
\citet{shirobokov_differentiating_2020} use learned differentiable surrogates to optimize parameters for generative models of small physics simulators.
This technique is similar to an iterative version of \name{} that continuously re-optimizes the surrogate around the current parameter table estimate.
\citet{louppe_adversarial_2017} propose optimizing non-differentiable physics simulators by formulating the joint optimization problem as adversarial variational optimization.
\citeauthor{louppe_adversarial_2017}'s technique is applicable in principle, though it has only been evaluated in small settings with a single parameter to learn.
\citet{grathwohl_backpropagation_2018} use learned differentiable surrogates to approximate the gradient of black-box or non-differentiable functions, in order to reduce the variance of gradient estimators of random variables.
While similar, \citeauthor{grathwohl_backpropagation_2018}'s surrogate optimization has a different objective: reducing the variance of other gradient estimators~\citep{reinforce}, rather than necessarily mimicking the black-box function.
\citet{she_neuzz_2019} use learned differentiable surrogates to approximate the branching behavior of real-world programs then find program inputs that trigger bugs in the program.
\citeauthor{she_neuzz_2019}'s surrogate does not learn the full input-output behavior of the program, only estimating which edges in the program graph are or are not taken.

\paragraph{CPU simulator surrogates}
\citet{ipek_exploring_2006} use neural networks to learn to predict the IPC of a cycle-accurate simulator given a set of design space parameters, to enable efficient design space exploration.
\citet{lee_illustrative_2007} use regression models to predict the performance and power usage of a CPU simulator, similarly enabling efficient design space exploration.
Neither \citeauthor{ipek_exploring_2006} nor \citeauthor{lee_illustrative_2007} then use the models to optimize the simulator to be more accurate; both also apply exhaustive or grid search to explore the parameter space, rather than using the gradient of the simulator~surrogate.

\paragraph{Differentiating arbitrary programs}
\citet{chaudhuri_smooth_2010} present a method to approximate numerical programs by executing programs probabilistically, similar to the idea of blurring an image.
This approach lets \citeauthor{chaudhuri_smooth_2010} apply gradient descent to parameters of arbitrary numerical programs.
However, the semantics presented by \citeauthor{chaudhuri_smooth_2010} only apply to a limited set of program constructs and do not easily extend to the set of program constructs exhibited by large-scale CPU simulators.

\begin{table*}
  \caption{Parameters learned for \exegesis{}.}
  \label{tab:exegesis-parameters}
  \centering
  \begin{tabularx}{\textwidth}{cccX}
    \toprule \textbf{Parameter} & \textbf{Count} & \textbf{Constraint} & \textbf{Description} \\ \midrule
    \latency & 1 per-instruction & Integer, $\geq 0$ & The number of cycles before destination operands of that instruction can be read from.  \\
    \portmap & 10 per-instruction & Integer, $\geq 0$ & The number of micro-ops dispatched to each port.
    \\ \bottomrule
  \end{tabularx}
\end{table*}

\begin{table}
\caption{Learning all parameters: error of \exegesis{} with the default and learned parameters.}
\label{tab:results-exegesis-all}
  \centering
    \begin{tabular}{p{12em}cc}
  \toprule
  \textbf{Architecture} \hfill \textbf{Predictor} & \textbf{Error} & \textbf{Kendall's Tau} \\
  \midrule    \textbf{Haswell} \hfill Default & $61.3\%$ & 0.7256 \\
                      \hfill \name{} & $44.1\%$ & 0.718 \\ \cmidrule{2-3}
                      \hfill Ithemal & $9.2\%$ & 0.854 \\
                      \hfill IACA & $17.1\%$ & 0.800 \\
                      \hfill OpenTuner & $115.6\%$ & 0.507 \\

  \bottomrule
\end{tabular}
\end{table}

\section{Conclusion}
CPU simulators are complex software artifacts that require significant measurement and manual tuning to set their parameters.
\NA{We present \name{}, a generic algorithm for learning parameters within non-differentiable programs, using only end-to-end supervision.}
Our results demonstrate that \name{} is able to learn the entire set of \nparams microarchitecture-specific parameters from scratch in \mca{}.
Looking beyond CPU simulation, \name{}’s approach offers the promise of a generic, scalable methodology to learn the parameters of programs using only input-output examples, potentially reducing many programming tasks to simply that of gathering data.

\section*{Acknowledgements}
We would like to thank Ben Sherman, Cambridge Yang, Eric Atkinson, Jesse Michel, Jonathan Frankle, Saman Amarasinghe, Ajay Brahmakshatriya, and the anonymous reviewers for their helpful comments and suggestions.
This work was supported in part by the National Science Foundation (NSF CCF-1918839, CCF-1533753), the Defense Advanced Research Projects Agency (DARPA Awards \#HR001118C0059 and \#FA8750-17-2-0126), and a Google Faculty Research Award, with cloud computing resources provided by the MIT Quest for Intelligence and the MIT-IBM Watson AI Lab.
Any opinions, findings, and conclusions or recommendations expressed in this material are those of the authors and do not necessarily reflect the views of~the~funding~agencies.

\appendices

\section{\exegesis{}}
\label[appendix]{sec:exegesis}

To evaluate that our implementation of \name{} (\Cref{sec:implementation}) is extensible to simulators other than \mca{}, we evaluate our implementation on \exegesis{}~\citep{exegesis}, learning all parameters that \exegesis{} reads from LLVM.
\exegesis{} is a simulator that uses many of the same parameters (from LLVM’s backend) as \mca{}, but uses a different model of the CPU, modeling the frontend and breaking up instructions into micro-ops and simulating the micro-ops individually rather than simulating instructions as a whole as \mca{} does.

\paragraph{Behavior}
\exegesis{}~\citep{exegesis} is also an out-of-order superscalar simulator exposing LLVM's instruction scheduling model.
\exegesis{} is only implemented for the x86 Haswell microarchitecture.
Similar to \mca{}, \exegesis{} also predicts timings of basic blocks, assuming that all data is in the L1 cache.
\exegesis{} primarily differs from \mca{} in two aspects: It models the front-end, and it decodes instructions into micro-ops before dispatch and execution.
\exegesis{} has the following pipeline:

\begin{itemize}
\item Instructions are fetched, parsed, and decoded into micro-ops (unlike \mca{}, \exegesis{} does model the frontend)
\item Registers are renamed, with an unlimited number of physical registers
\item Micro-ops are dispatched out-of-order once dependencies are available
\item Micro-ops are executed on execution ports
\item Instructions are retired once all micro-ops in an instruction have~been~executed
\end{itemize}

\paragraph{Parameters}
We learn the parameters specified in \Cref{tab:exegesis-parameters}.
We again assume that there are 10 execution ports available to dispatch for all microarchitectures and do not learn to dispatch to port groups.
All other hyperparameters are identical to those described in \Cref{sec:methodology}.

\paragraph{Results}
Table~\ref{tab:results-exegesis-all} presents the average error and correlation of \exegesis{} with the default parameters, \exegesis{} with the learned parameters, Ithemal trained on the dataset as a lower bound,
and the OpenTuner~\citep{ansel_opentuner_2014} baseline.
By learning the parameters that \exegesis{} reads from LLVM, we reduce \exegesis{}'s error from $61.3\%$ to $44.1\%$.

\bibliographystyle{IEEEtranN}
\IEEEtriggeratref{15}
\IEEEtriggercmd{\balance}
\bibliography{references}
\end{document}